\documentclass[conference]{IEEEtran}
\IEEEoverridecommandlockouts
\usepackage{cite}
\usepackage{amsmath,amssymb,amsfonts}
\usepackage{algorithmic}
\usepackage{graphicx}
\usepackage{subfigure}
\usepackage{textcomp}
\usepackage{multirow}
\usepackage{xcolor}
\usepackage{array}
\def\BibTeX{{\rm B\kern-.05em{\sc i\kern-.025em b}\kern-.08em
    T\kern-.1667em\lower.7ex\hbox{E}\kern-.125emX}}
\begin{document}

\title{Latent Preserving Generative Adversarial Network for Imbalance classification}

\author{\IEEEauthorblockN{Tanmoy Dam}
\IEEEauthorblockA{\textit{SEIT} \\
\textit{University of New South Wales}\\
Canberra, Australia}
\and
\IEEEauthorblockN{Md Meftahul Ferdaus}
\IEEEauthorblockA{\textit{ATMRI} \\
\textit{Nanyang Technological University}\\
Singapore}
\and
\IEEEauthorblockN{Mahardhika Pratama, ~\IEEEmembership{Senior Member,~IEEE}}
\IEEEauthorblockA{\textit{STEM} \\
\textit{University of South Australia}\\
Adelaide, Australia}
\and
\IEEEauthorblockN{Sreenatha G. Anavatti}
\IEEEauthorblockA{\textit{SEIT} \\
\textit{University of New South Wales }\\
Canberra, Australia}
\and
\IEEEauthorblockN{Senthilnath Jayavelu, ~\IEEEmembership{Senior Member IEEE}}
\IEEEauthorblockA{\textit{Institute for Infocomm Research} \\
\textit{A*STAR}\\
Singapore}
\and
\IEEEauthorblockN{Hussein A. Abbass, ~\IEEEmembership{Fellow,~IEEE}}
\IEEEauthorblockA{\textit{SEIT} \\
\textit{University of New South Wales}\\
Canberra, Australia}
}

\maketitle

\begin{abstract}
Many real-world classification problems have imbalanced frequency of class labels; a well-known issue known as the ``class imbalance'' problem. Classic classification algorithms tend to be biased towards the majority class, leaving the classifier vulnerable to misclassification of the minority class. While the literature is rich with methods to fix this problem, as the dimensionality of the problem increases, many of these methods do not scale-up and the cost of running them become prohibitive. In this paper, we present an end-to-end deep generative classifier. We propose a domain-constraint autoencoder to preserve the latent-space as prior for a generator, which is then used to play an adversarial game with two other deep networks, a discriminator and a classifier. Extensive experiments are carried out on three different multi-class imbalanced problems and a comparison with state-of-the-art methods. Experimental results confirmed the superiority of our method over popular algorithms in handling high-dimensional imbalanced classification problems. Our code is available on https://github.com/TanmDL/SLPPL-GAN.
\end{abstract}


\section{Introduction}
Class imbalance classification is an old-standing problem. Common methods to address the problem include cost-sensitive classification, undersampling, and oversampling techniques~\cite{branco2016survey}. The former requires a domain expert to transform precision and recall rates into a utility function. Undersampling the majority class could lead to information loss. Oversampling the minority class has been commonly used \cite{chawla2002smote,he2008adasyn} , but raises the main challenge solved in this paper: how to generate new meaningful samples.

A deep oversampling framework (DOS) was proposed in \cite{ando2017deep} to fulfill the end-to-end requirement of deep learning algorithms. A limitation of the DOS lies in its dependency on the class-wise neighborhood sizes, which are determined by costly parameter tuning. The generative adversarial network (GAN) \cite{goodfellow2014generative} has gained popularity due to its unique capability in generating synthetic but realistic samples. GAN relies on random noises that may yield a highly entangled process and disruptions of feature orientations. A two-stage framework called BAGAN \cite{mariani2018bagan} is proposed by combining autoencoder (AE) with conditional GAN (cGAN) \cite{mirza2014conditional}. The latent code learnt via AE is fed to the cGAN to replace random noises. Because of its power in generating realistic samples, GAN is applied to oversample minority class(es) \cite{douzas2018effective}. This approach often leads to boundary distortion as witnessed in \cite{mullick2019generative,santurkar2018classification}. This problem inspires the development of a discriminative feature-based sampling (DFBS) method, where the main goal is to produce discriminative latent features achieved via the use of the triplet loss to learn AE. This approach can cause intra-class instances to stay together, while inter-class instances are pushed apart.

Instances generated by their method are likely to be close to the boundaries of the minority class(es), calling for a reliable classifier \cite{blundell2015weight,srivastava2017veegan}. A deep generative classifier (DGC) is developed in \cite{wang2020deep} for solving unstable prediction problems in the imbalance classification problem. A model is perturbed by replacing the fixed values of latent variables with a probability distribution over possible values, whereas the data are perturbed by generating feature/label. To mitigate the majority class influence on the classifier, probabilistic latent codes are over-sampled at different fractions rates. However, the learned probabilistic latent codes can\textquoteright t guarantee well separation at the encoded manifold that may adversely influence the classifier\textquoteright s performance.

Generative adversarial minority oversampling (GAMO) is proposed in \cite{mullick2019generative}. A mixture of convex generators is proposed to mitigate the generation of majority class samples at the generator side. The mixture of generators forcefully generates samples within a specific minority class distribution by utilizing real instances from that minority class. However, their generators are limited by generating class instances instead of generating the real data distribution. To determine class instances from the real data distribution, an adversarial game is played between the mixture of generators, a discriminator and a classifier. True data distributions could be far from such a convex hull of minority class(es) leading to the generation of instances with low information or overlapped instances.

In deep generative classifiers, cross entropy loss \cite{wang2020deep} or mean-squared loss \cite{mullick2019generative} are commonly used to update the classifier. These loss functions may create hard partition(s) between majority and minority classes, where the decision boundary is influenced by the majority class samples leading to over-fitting. To mitigate the influence of majority class samples on the decision boundary, a classifier is updated twice in \cite{wang2020deep, mullick2019generative} based on concepts borrowed from \cite{statistics1986approach,koh2017understanding}. First, it is updated through the real feature distribution. Afterward, generated minority class samples are used to update the classifier. A similar approach is also followed in this paper. In addition, mixtures of minority class-specific generators are replaced with a single generator only. This is done by deploying a domain-constrained AE to learn the class-specific latent code, preserved and used as a prior for the generative network. Besides, in GAMO, the minority samples are generated only from the feature space while ignoring the possibility of an adversarial data space oversampling approach. Thus, the only three-player strategy of adversarial minority oversampling of GAMO is extended here to two different three-player strategies: 1) adversarial minority oversampling (AMO); 2) adversarial data space oversampling (ADSO).

The main contributions of this paper are summarized:
\begin{itemize}
  \item We define a joint learning framework to preserve the latent space in a low dimensional manifold by utilizing a supervised autoencoder (AE). The learned latent space works as a prior for a single generator, which engages in an adversarial game with a classifier and a discriminator. Replacement of minority class(es)-convex generators with only one generator makes our proposed three player adversarial architecture more scalable than GAMO \cite{mullick2019generative}.   
  \item We leveraged all two possible strategies among three players game to improve the classifier\textquoteright s performance. First strategy is related to ADSO and the last strategy is based on AMO.
  \item A set of experiments have been carried out under different imbalance ratios, where the experimental results support that an ADSO-based classifier performs better than the AMO-based approach and state-of-the-art baselines, and it does so with significant margins.  
\end{itemize}

\section{Problem Formulations}
Let us assume a set of $N$ samples $\{x_i,y_i\}_{i=1}^N\in (X_{org}, Y_{org})$, a multi-class imbalance dataset distributed over $C$ classes. Here, ${\bf x}_i \in R^d$ is an $i$-th input image with its corresponding target class ${\bf y}_i$. Without loss of generalisation,  the class distribution follows $p_{maj} \ge p_2\ge... p_l\ge ...\ge p_{min}$, $N = \sum_{l=c}^{C} p_{l}$, in which $p_{maj}$ and $p_{min}$ denote the majority class and the minority class, respectively. The relationship between the majority class and minority class(es) is set as $p_{maj} \geq 50*p_{min}$ following the problem setting of \cite{dam2021does, mullick2019generative, dam2020mixture}.  The main objective is to design a deep neural network estimating the underlying data distribution of $C$ classes, thereby producing robust decision boundaries.

 The network structure of our approach consists of three parts: supervised latent preserving prior learning (SLPPL), ADSO and AMO. The adversarial game is played among preserved latent prior generator($G_{\omega}$), a discriminator($Dis_{\xi}$) and a classifier($Q_{\varrho}$) to improve the classification performance. In SLPPL, the $Enc_{\theta}$ encodes the original class instances ($x_i \in X_{org}$) into a lower-dimensional latent space ($z_i \in Z$) and $Dec_{\phi}$ takes the encoded latent space ($z_i$) to produce the reconstructed original data ($\hat x_i \in X_{org}$). The $Enc_{\theta}$ learns class distribution ($Enc_{\theta}(y_i|x_i)$) for $i$-th class instance. After learning the latent space ($z_i$) from data directly, a Gaussian multivariate normal distribution is constructed as a prior for $G_{\omega}$. In the adversarial game, we introduce mainly two strategies, ADSO and AMO, among the pretrained prior $G_{\omega}$, a $Dis_{\xi}$ and a $Q_{\varrho}$.

\section{Our Approach}

\subsection{SLPPL}
High dimensional data always retain their characteristics in a low dimensional encoded manifold that inspired us to design a latent prior for each class distribution \cite{ghosh2019variational}. A low dimensional manifold learning  approach attains optimum performance by freely moving the latent space when the distribution of data is uniform in nature \cite{ghosh2019variational}. If data are not distributed uniformly, the majority class-driven latent space always dominates minority classes. To mitigate a biased prediction problem favouring only the majority class, we utilize jointly ($X_{org}, Y_{org}$) a class distribution learning approach as well as a reconstruction learning approach to deliver the latent space bounded as much as possible. The bounded latent space is obtained by considering label information in SLPPL under the deterministic autoencoder (AE) framework. Unlike, arbitrarily chosen prior based on VAE \cite{kingma2013auto}, this approach reduces the stochasticity in the latent space. Finally, the obtained latent space is derived as a significant smooth manifold. The AE objective is to minimize the reconstruction loss ($L_{rec}$) between input data ($X_{org}$) and decoded output  $(Dec_{\phi}(Enc_{\theta}(X_{org})))$:

\begin{multline}
L_{rec}(X_{org},Dec_{\phi}(Enc_{\theta}(X_{org})))= \\ \mathbb{E}_{x_i\in X_{org}}||x_{i}-Dec_{\phi}(Enc_{\theta}(x_{i}))||_{2}
\end{multline}

Similarly, for the bounded latent space, the encoder network ($Enc_{\theta}$) estimates the class distribution ($Y_{org}$).
\begin{equation}
    L_{bce}(Y_{org},(Enc_{\theta}(X_{org})))=\mathbb{E}_{y_i\in Y_{org}}y_{i}log(Enc_{\theta}(x_{i}))
\end{equation}
Finally, merging two losses, the final loss function for SLPPL can be minimized with respect to two networks parameter.
\begin{equation}\label{loss}
    \min_{\theta,\phi }L_{SLPPL}= L_{rec} + L_{bce} 
\end{equation}
The proposed SLPPL is able to preserve stable network parameters $(\phi , \theta)$, and to maintain stable class distributions in the low-dimensional manifold where \eqref{loss} is solved using the ADAM optimiser \cite{dam2021does}. The $i$-th sample is easily encoded in the latent space afterward as $z_i \in Z= Enc_{\theta}(x_i)$ and preserves a multivariate normal distribution (MND) \cite{ghosh2019variational} for the $i$-th class. The MND definition for the $i$-th class is  $z_i \in \mathcal{N}(\mu_i,\,\sigma_i^{2})$ where $\mu_i \in R^q $ and $\sigma_i \in  R^{q \times q}$  are the mean and variance of the latent space respectively. The prior $(z_i)$ preserving latent space is applied to improve image generation through adversarial game between $Q_{\varrho}$ and $Dis_{\xi}$ through two adversarial strategies: ADSO and AMO. 
In both the oversampling cases, the $G_{\omega}$ network structure is same as $Dec_{\phi}$. Hence, the initialisation of $G_{\omega}$ network weights is taken from pretrained $Dec_{\phi}$, whereas the structure of $Dis_{\xi}$ and $Q_{\varrho}$ are almost similar to $Enc_{\theta}$ network but  the last layer of $Dis_{\xi}$ gives a single output. Besides, the learned feature layers of $Enc_{\theta}$ are used to initialise the weights of both the $Dis_{\xi}$ and $Q_{\varrho}$. However, to reduce the over-fitting effect at $Q_{\varrho}$, we apply dropout after activation function \cite{hinton2012improving}.

\subsection{ ADSO \& AMO}

In ADSO, the minority class(es) is repeated to form a balanced data distribution ($X_{bal}, Y_{bal}$) in the data space. An adversarial game is played among $G_{\omega}$, $Dis_{\xi}$ and $Q_{\varrho}$ afterward, where the $G_{\omega}$ network is updated by fooling only the discriminator $Dis_{\xi}$ but by favouring the classifier $Q_{\varrho}$. The generator network $G_{\omega}$ aims to generate samples to be classified by $Q_{\varrho}$ as the same class, and thus, the generated samples assign real scores while updating the generator $G_{\omega}$ through the classifier $Q_{\varrho}$. The discriminator $Dis_{\xi}$ enforces the generator $G_{\omega}$ to follow the real data distribution. While updating the classifier $Q_{\varrho}$ through the generator $G_{\omega}$, generated samples assign a fake score to the classifier $Q_{\varrho}$. In other words, the classifier $Q_{\varrho}$ assigns a high probability in such a way that the generated samples are classified as other classes. We can formulate the following optimisation problem for three players adversarial game among $G_{\omega}$, $Dis_{\xi}$ and $Q_{\varrho}$:
\begin{equation}
    \min_{\omega, \varrho}\max_{\xi} L_{ADSO}(G_{\omega},Dis_{\xi},Q_{\varrho}) 
\end{equation}
The total loss can be expressed as $L_{ADSO}= L_{ADSO}^{G_{\omega}} + L_{ADSO}^{Dis_{\xi}} + L_{ADSO}^{Q_{\varrho}}$, where
\begin{multline}
L_{ADSO}^{G_{\omega}} = \mathbb{E}_{G(z_i) \in Z} (f(1-Dis_{\xi}(G(z_i))) + \\ \mathbb{E}_{G(z_i) \in Z}  (y_{i}\log(Q_{\varrho}(G_{\omega}(z_i)) )
\end{multline}

\begin{multline}
L_{ADSO}^{Q_{\varrho}} = \mathbb{E}_{y_i\in Y_{bal}}y_{i}log(Q_{\varrho}(x_{i})) + \\
\mathbb{E}_{G(z_i) \in Z}  (y_{i}\log(1- Q_{\varrho}(G_{\omega}(z_i)) )
\end{multline}

\begin{multline}
L_{ADSO}^{Dis_{\xi}} = \mathbb{E}_{x_i \in X_{bal}} (f(Dis_{\xi}(x_i)) + \\ \mathbb{E}_{G(z_i) \in Z}  f(1-Dis_{\xi}(G_{\omega}(z_i)))
\end{multline}

In AMO, generator network $G_{\omega}$ aims to generate samples to be classified by $Q_{\varrho}$ as the same class, and thus, the generated samples assign real scores while updating the generator $G_{\omega}$ through the classifier $Q_{\varrho}$. To mitigate, majority class biases at $Q_{\varrho}$,  is updated through the real samples as well as minority-class generated samples. $Dis_{\xi}$ network forces $G_{\omega}$ to learn the real-data distributions. The following optimisation can be formulated by playing three players game among $G_{\omega}$, $Dis_{\xi}$ and $Q_{\varrho}$:
\begin{equation}
    \min_{\omega, \varrho}\max_{\xi} L_{AMO}(G_{\omega},Dis_{\xi},Q_{\varrho}) 
\end{equation}
The total loss can be expressed as $L_{AMO}= L_{AMO}^{G_{\omega}} + L_{AMO}^{Dis_{\xi}} + L_{AMO}^{Q_{\varrho}}$, where

\begin{multline}
L_{AMO}^{G_{\omega}} = \mathbb{E}_{G(z_i) \in Z} (f(1-Dis_{\xi}(G(z_i))) + \\ \mathbb{E}_{G(z_i) \in Z}  (y_{i}\log(1-Q_{\varrho}(G_{\omega}(z_i)) )
\end{multline}

\begin{multline}
L_{AMO}^{Q_{\varrho}} = \mathbb{E}_{y_i\in Y_{org}}y_{i}log(Q_{\varrho}(x_{i})) + \\
\mathbb{E}_{G(z_j) \setminus p_{maj} \in Z}  (y_{j}\log(Q_{\varrho}(G_{\omega}(z_j)) )
\end{multline}

\begin{multline}
L_{AMO}^{Dis_{\xi}} = \mathbb{E}_{x_i \in X_{org}} (f(Dis_{\xi}(x_i)) + \\ \mathbb{E}_{G(z_i) \in Z}  f(1-Dis_{\xi}(G_{\omega}(z_i)))
\end{multline}

Cross-entropy (CE) loss and complementary CE (CCE) loss are represented by the expressions $log Q(.)$ and $log(1-Q(.))$, respectively. For both the cases (AMO, ADSO), the functional operator $f(.)$  selects different GANs types. For vanilla \cite{goodfellow2014generative} and Wasserstein GANs (WGANs)\cite{mescheder2018training}, $(f)$ is represented as $f(x)=\log x$ and $f(x)=x$ respectively. We follows WGAN's zero center ($0$)-gradient penalty (0-gp) for all the GAN strategies \cite{mescheder2018training}.

\begin{table*}
\centering
\caption{The detailed description of Experimental datasets}\label{tab:data description}
\begin{tabular}{>{\raggedright}p{1.3cm}|>{\raggedright}p{1.6cm}|>{\raggedright}p{0.8cm}|>{\raggedright}p{1.0cm}|>{\raggedright}p{3.7cm}|>{\raggedright}p{3.6cm}}
\hline
Datasets & Data Dimensions & IR & Classes & Training Set & Testing Set\tabularnewline
\hline 
MNIST & $28\times28\times1$  & $100$ & $10$ & $\begin{array}{c}
[4000,2000,1000,750,\\
500,350,200,100,60,40]
\end{array}$ & $\begin{array}{c}
[980,1135,1032,1010,982,\\
892,985,1028,974,1009]
\end{array}$\tabularnewline
\hline 
Fashion-MNIST & $28\times28\times1$ & $100$ & $10$ & $\begin{array}{c}
[4000,2000,1000,750,\\
500,350,200,100,60,40]
\end{array}$ & $\begin{array}{c}
[1000,1000,1000,1000,1000,\\
1000,1000,1000,1000,1000]
\end{array}$\tabularnewline
\hline
CelebA & $64\times64\times3$ & $100$ & $5$ & $\begin{array}{c}
[15000,1500,750,300,150]
\end{array}$ & $\begin{array}{c}
[2660,5422,412,3428,535]
\end{array}$\tabularnewline
\hline 
\end{tabular}
\end{table*}

\begin{table*}
\caption{Overall classification performance on various datasets}
\label{tab:acsa_F_G}
\centering{}%
\begin{tabular}{c|c|c|c|c|c|c|c|c|c}
\hline 
\multirow{2}{*}{Methods} & \multicolumn{3}{c|}{MNIST} & \multicolumn{3}{c|}{FMNIST} & \multicolumn{3}{c}{CelebA}\tabularnewline
\cline{2-10} \cline{3-10} \cline{4-10} \cline{5-10} \cline{6-10} \cline{7-10} \cline{8-10} \cline{9-10} \cline{10-10} 
 & $ACSA$ & $F_{macro}$ & $G_{macro}$ & $ACSA$ & $F_{macro}$ & $G_{macro}$ & $ACSA$ & $F_{macro}$ & $G_{macro}$\tabularnewline
\hline 
BAGAN & 0.8848 & 0.8785 & 0.9295 & 0.7814 & 0.7610 & 0.8546 & 0.5972 & 0.5152 & 0.6554\tabularnewline
\hline 
DFBS & 0.7812 & 0.7838 & 0.8683 & 0.5135 & 0.4620 & 0.6382 & 0.2109 & 0.1335 & 0.2664\tabularnewline
\hline 
GAMO & 0.8826 & 0.8794 & 0.9308 & 0.7929 & 0.7880 & 0.8740 & 0.6409 & 0.5903 & 0.7472\tabularnewline
\hline 
BayesCNN & 0.9158 & 0.9141 & 0.9512 & 0.7934 & 0.7835 & 0.8701 & 0.5517 & 0.4936 & 0.6534\tabularnewline
\hline 
mmDGMs & 0.9066 & 0.9039 & 0.9449 & 0.8091 & 0.7984 & 0.8796 & 0.3760 & 0.0618 & 0.3754\tabularnewline
\hline 
DGC & 0.9480 & 0.9474 & 0.9704 & 0.8364 & 0.8314 & 0.9010 & 0.6755 & 0.6454 & 0.7779\tabularnewline
\hline 
DSO+$Q_\varrho$ & 0.9339 & 0.9325 & 0.9619 & 0.8450 & 0.8436 & 0.9089 & 0.7078 & 0.6663 & 0.7997\tabularnewline
\hline 
AMO & 0.9403 & 0.9386 & 0.9656 & 0.8428 & 0.8378 & 0.9046 & 0.6702 & 0.6210 & 0.7628\tabularnewline
\hline 
ADSO & \textbf{0.9613} & \textbf{0.9609} & \textbf{0.9781} & \textbf{0.8675} & \textbf{0.8648} & \textbf{0.9221} & \textbf{0.7595} & \textbf{0.7217} & \textbf{0.8359}\tabularnewline
\hline 
\end{tabular}
\end{table*}

\begin{table}
\caption{Classification performance on the largest ($P_{maj}$) and smallest
class ($R_{min}$)}\label{tab:recall_pre}

\centering{}%
\begin{tabular}{>{\raggedright}p{1.6cm}|>{\centering}p{0.7cm}|>{\centering}p{0.7cm}|>{\centering}p{0.7cm}|>{\centering}p{0.7cm}|>{\centering}p{0.7cm}|>{\centering}p{0.7cm}}
\hline 
\multirow{2}{1.2cm}{Methods } & \multicolumn{2}{c|}{MNIST} & \multicolumn{2}{c|}{FMNIST} & \multicolumn{2}{c}{CelebA}\tabularnewline
\cline{2-7} \cline{3-7} \cline{4-7} \cline{5-7} \cline{6-7} \cline{7-7} 
 & $R_{min}$ & $P_{maj}$ & $R_{min}$ & $P_{maj}$ & $R_{min}$ & \multicolumn{1}{c}{$P_{maj}$}\tabularnewline
\hline 
BAGAN & 0.5354 & 0.8541 & 0.7306 & 0.5709 & 0.0192 & 0.5064\tabularnewline
\hline 
DFBS & 0.5946 & 0.5118 & 0.4412 & 0.3395 & 0.0522 & 0.2174\tabularnewline
\hline 
GAMO & 0.6394 & 0.8812 & 0.7928 & 0.6165 & 0.2302 & 0.6687\tabularnewline
\hline 
BayesCNN & 0.7578 & 0.8896 & 0.8474 & 0.6022 & 0.1063 & 0.5225\tabularnewline
\hline 
mmDGMs & 0.6525 & 0.8459 & 0.8160 & 0.5942 & 0.0006 & 0.4110\tabularnewline
\hline 
DGC & 0.8276 & 0.9270 & 0.8864 & 0.6900 & 0.2987 & 0.7603\tabularnewline
\hline 
DSO+$Q_\varrho$ & 0.7393 & 0.8761 & 0.9130 & 0.7598 & 0.2783 & 0.7483\tabularnewline
\hline 
AMO & 0.7363 & 0.9055 & 0.9170 & 0.6752 & 0.2405 & 0.5785\tabularnewline
\hline 
ADSO & \textbf{0.8840} & \textbf{0.9348} & \textbf{0.9510} & \textbf{0.7885} & \textbf{0.3223} & \textbf{0.8132}\tabularnewline
\hline 
\end{tabular}
\end{table}

\section{Experiments}

\subsection{Datasets}
 Two single-channel (MNIST \cite{lecun1998mnist}, and Fashion-MNIST \cite{xiao2017fashion}) and a three-channel (CelebA \cite{liu2015deep}) image sets are used here. Properties of these datasets are tabulated in Table \ref{tab:data description} where IR indicates the imbalance ratio.

\subsection{Evaluation metrics and baselines}
Five metrics are used here to measure the imbalance classification performance: 1) average class specific accuracy (ACSA); 2) macro-averaged F-measure ($F_{macro}$); 3) macro-averaged geometric mean ($G_{mean}$); 4) precision of majority class ($P_{maj}$); and 5) recall of minority class ($R_{min}$).

Our proposed method is compared against seven different baselines namely BAGAN \cite{mariani2018bagan}, DFBS \cite{liu2018deep}, GAMO \cite{mullick2019generative}, mmDGMs \cite{li2017max}, BayesCNN \cite{shridhar2019comprehensive}, DGC \cite{wang2020deep}, and data space oversampling (DSO) plus baseline $Q_{\varrho}$. Results of all above-mentioned baselines are adopted from \cite{wang2020deep}. For our two proposed models, the size of latent space ($z_i \in R^q $) is set to $q=64$ for the MNIST and Fashion-MNIST datasets \cite{wang2020deep}, assigned as $q =128$ for the CelebA dataset \cite{wang2020deep}. 

\subsection{Numerical Results}

The overall classification performances in terms of average of ACSA, $F_{macro}$, and $G_{macro}$ are reported in Table \ref{tab:acsa_F_G}. The ADSO-based strategy we suggest is the one that performs the best overall for handling imbalance classification. In contrast with the ADSO, some limitations of baselines for getting comparative poorer performance are as follows: DFBS can not create sufficient margins among classes. GAMO utilizes computationally expensive MSE loss that requires real samples in each generator. The mode collapse problem may appear in BAGAN due to the initialization mechanism of subsequent GAN. Though Bayes CNN has adopted a model perturbation strategy, very few instances in minority class(es) may not be sufficient to train the complicated model. mmDGMs determines class boundaries by adopting the discriminative classifier, limiting their performance in imbalanced datasets. The perturbation mechanism of both data and model supports the DGC to outperform the above-mentioned baselines. However, well separation at the encoded manifold can not be guaranteed by the learned probabilistic latent codes. A high recall on minority class is expected from the $Q_{\varrho}$ while maintaining a high precision on majority class. Recall of the smallest class ($R_{min}$) and precision of the largest class ($P_{maj}$) are listed in Table \ref{tab:recall_pre}. For Fashion-MNIST, it is observed that the performance of all six baselines have improved in the minority class but has not been significant in the majority class. These results confirm that class boundaries are not determined clearly by these existing methods. In the three-channel CelebA image dataset, poor performance is seen in both majority and minority classes from the first five baselines. Since the learned probabilistic latent in DGC does not guarantee well separation at the encoded manifold. Similar phenomena are also observed in minority generative samples for our proposed AMO method because of the majority class influences at $Q_{\varrho} $.  Even if we preserve the prior for the $G_{\omega}$, $Q_{\varrho} $ has influenced it to generate majority classes, i.e., AMO can't beat ADSO. In contrast, under three-player GAN settings, the proposed ADSO-based $G_{\omega}$ tries to fool itself by generating a subset of each class sample that is relevant to $Q_{\varrho}$.







\section{Conclusions}
We propose a latent preserving based deep generative model for handling imbalanced classification problems. The class constraints AE is used to preserve the latent space, utilized as a prior for $G_{\omega}$. By playing two adversarial games among the latent space preserved across $G_{\omega}$, a $Dis_{\xi}$, and a $Q_\varrho$, improvement in the $Q_\varrho$'s performance is witnessed. From the experimental results on all three datasets, our ADSO-based strategy performed better than all the baselines. Our future work includes real-world applications like semiconductor fault detection, medical diagnosis, etc.

\section*{Acknowledgements}
S. Jayavelu and Md M. Ferdaus acknowledges funding from the Accelerated Materials Development for Manufacturing Program at A*STAR via the AME Programmatic Fund by the Agency for Science, Technology and Research under Grant No. A1898b0043. T. Dam acknowledges UIPA funding from UNSW Canberra. 

\bibliographystyle{IEEEtran}
\bibliography{ref_ICIP22}

\end{document}